\documentclass[journal]{IEEEtran}
\usepackage[T1]{fontenc}
\usepackage{amsmath,amssymb,bm}
\usepackage{graphicx,booktabs,array,multirow}
\usepackage{cite}
\usepackage{xcolor}
\usepackage{svg}
\usepackage[table]{xcolor}
\usepackage[hidelinks]{hyperref}

\usepackage{placeins}

\hypersetup{
    pdftitle={CP-JEPA: Context-Predictive Low-Rank Adaptation for Face Presentation Attack Detection},
    pdfauthor={Aminu Yusuf; Shakeel Muhammad Ibrahim; Sulaimon Oyeniyi Adebayo; Muhammad Sohail Ibrahim; Muhammad Shahid Jabbar; Shujaat Khan}
}

\graphicspath{{figures/}}

\newcommand{\CE}{\mathcal{L}_{\mathrm{CE}}}
\newcommand{\LP}{\mathcal{L}_{\mathrm{pred}}}
\newcommand{\LM}{\mathcal{L}_{\mathrm{mask}}}
\newcommand{\LC}{\mathcal{L}_{\mathrm{cons}}}
\newcommand{\R}{\mathbb{R}}
\newcommand{\method}{CP-JEPA}

\usepackage[numbers,sort&compress]{natbib}
\usepackage{etoolbox}

\let\OriginalCite\cite

\newcommand{\CiteWithYear}[1]{%
    \OriginalCite{#1} (\citeyear{#1})%
}

\newcommand{\EnableTableCitationYears}{%
    \AtBeginEnvironment{tabular}{%
        \let\cite\CiteWithYear
    }%
}

\EnableTableCitationYears

\begin{document}
\title{Context-Predictive Low-Rank Adaptation for Face Presentation Attack Detection}
\author{Aminu Yusuf, Shakeel Muhammad Ibrahim, Sulaimon Oyeniyi Adebayo, Muhammad Sohail Ibrahim,\\ Muhammad Shahid Jabbar, and Shujaat Khan%
\thanks{Aminu Yusuf, Sulaimon Oyeniyi Adebayo and Shujaat Khan are with the Department of Computer Engineering, College of Computing and Mathematics, King Fahd University of Petroleum \& Minerals, Dhahran 31261, Saudi Arabia.}
\thanks{Shakeel Muhammad Ibrahim is with Geomexsoft Co. Ltd., Chuncheon-si, Republic of Korea.}
\thanks{Muhammad Sohail Ibrahim is with the Interdisciplinary Research Center for Intelligent Secure Systems, King Fahd University of Petroleum \& Minerals, Dhahran 31261, Saudi Arabia.}
\thanks{Muhammad Shahid Jabbar and Shujaat Khan are with the SDAIA-KFUPM Joint Research Center for Artificial Intelligence, King Fahd University of Petroleum \& Minerals, Dhahran 31261, Saudi Arabia. Corresponding author: Shujaat Khan (e-mail: shujaat.khan@kfupm.edu.sa).}}
\maketitle

\begin{abstract}
Face presentation attack detection requires representations that distinguish physical spoofing cues across changes in acquisition devices, subjects, and attack media. Large visual encoders offer transferable representations, but conventional binary fine-tuning supervises their final decisions without explicitly constraining how local facial evidence is represented under partial observation. We present CP-JEPA, a context-predictive adaptation framework that combines low-rank tuning of a pretrained CLIP image encoder with a joint-embedding predictive objective. A student predicts complete-image teacher features at masked spatial locations, while unmasked classification, masked classification, and prediction consistency retain the connection to the live/spoof task. An exponential-moving-average teacher supplies evolving latent targets. The auxiliary components are used only during training; inference requires only the LoRA-adapted CLIP visual encoder and binary classifier. {We evaluate CP-JEPA under single-source and multi-source OCIM transfer, additional CelebA-Spoof training, transfer to CelebA-Spoof, two-source training, and unseen-attack evaluation on SiW-Mv2.} On the standard OCIM benchmark, CP-JEPA achieves 3.74\% mean HTER and 98.73\% mean AUC. {Single-source evaluation achieves 11.36\% mean HTER across twelve source--target combinations.} With CelebA-Spoof added to the three OCIM training sources, CP-JEPA reduces mean HTER from 3.45\% to 2.35\% and increases mean AUC from 98.68\% to 99.32\%. On SiW-Mv2, it achieves 3.36\% mean HTER across the held-out attack categories. Controlled loss ablations examine the contributions of latent prediction, masked classification, and prediction consistency. The results show that context-predictive adaptation can improve cross-dataset PAD performance, although the gains vary with the training sources and evaluation setting. The implementation is available at the \url{https://github.com/aminuyusufmr/cp_jepa} repository.
\end{abstract}


\section{Introduction}
\IEEEPARstart{F}{ace} recognition is increasingly deployed through commodity cameras, making presentation attack detection (PAD) an essential component of biometric authentication. Printed photographs and replayed videos can present the appearance of a legitimate user without the physical presence of that user. A PAD system must therefore distinguish a genuine face presentation from an attack, even when both depict the same identity. The difficulty increases when deployment changes the sensor, illumination, display, printing process, or capture environment. A cue that separates classes within one dataset can reflect its acquisition process as much as the underlying presentation attack.

Research has addressed this challenge through task-specific local operators, domain-aware representation learning, multimodal fusion, and large pretrained models. Central-difference networks emphasize local intensity structure~\cite{yu2020searching}; domain-generalization methods constrain the geometry of bona fide and attack representations~\cite{jia2020single,sun2023safas,zhou2023iadg}. Recent studies extend the design space through decomposition--fusion representations~\cite{dfnet2026}, domain-adversarial cross-attention~\cite{grdx2026}, and multimodal bidirectional fusion~\cite{mbca2026}. Pretrained visual encoders such as CLIP can also be adapted for PAD using low-rank updates~\cite{ozgur2025foundpad}.

Binary classification during low-rank adaptation encourages features that distinguish bona fide presentations from attacks but does not explicitly supervise the prediction of masked spatial representations. Masked feature prediction uses visible context to estimate teacher representations at masked locations. This distinction motivates combining global classification with local feature prediction, without requiring pixel reconstruction, depth annotations, physiological measurements, or an additional inference network.

We develop CP-JEPA to combine supervised CLIP adaptation with masked feature prediction. After an initial cross-entropy stage, the student learns to predict spatial features from an EMA teacher that receives the complete image. The student retains its native classification pathway, with additional losses for masked classification and consistency between masked and unmasked predictions. We adapt the joint-embedding predictive principle~\cite{ijepa2023} to low-rank PAD training and examine the contribution of each auxiliary objective separately.

The contributions are as follows:
\begin{itemize}
    \item We formulate a context-predictive low-rank adaptation framework that couples native CLIP classification with dense EMA-teacher feature prediction under input masking.

    \item We incorporate masked classification and prediction consistency alongside latent prediction during supervised CLIP-LoRA adaptation. The teacher and predictor are used only during training, leaving the student classification architecture unchanged at inference.

    \item {We evaluate CP-JEPA on cross-dataset and unseen-attack protocols, including single-source and multi-source OCIM transfer, additional CelebA-Spoof training, transfer to CelebA-Spoof, two-source training, and SiW-Mv2.} In the additional-source setting, mean HTER improves by 1.10 percentage points and mean AUC by 0.64 points over the CLIP-LoRA baseline.
\end{itemize}

\section{Related Work}
\label{sec:related}
\subsection{Representation Learning for Cross-Dataset PAD}

Early deep PAD models combined binary classification with local or auxiliary supervision. Central-difference convolution emphasizes fine spatial variations through intensity and gradient information~\cite{yu2020searching}. Domain-generalization methods subsequently examined the relationship between class discrimination and domain variation. Single-side domain generalization encourages compact bona fide representations while accommodating heterogeneous attacks~\cite{jia2020single}. SA-FAS considers class separability and domain alignment jointly~\cite{sun2023safas}, while instance-aware domain generalization addresses sample-dependent variability~\cite{zhou2023iadg}. DRDN combines residual distillation of semantic and low-frequency illumination information with attention-based contrastive learning~\cite{he2023dynamic}.

Recent works further diversify the available supervision. DFNet decomposes and fuses facial representations to support generalized and interpretable PAD~\cite{dfnet2026}. GR-DXNet combines multimodal cross-attention with adversarial domain suppression~\cite{grdx2026}. FAS-MBCA uses multiple paths and bidirectional cross-attention for multimodal PAD~\cite{mbca2026}. PCGattnNet uses dynamic graph attention on smartphone-acquired 3D facial point clouds to detect presentation attacks, with evaluation across unseen attack instruments and acquisition devices~\cite{ramachandra2025pcgattnnet}. Modality- and domain-aware adaptivity extends vision--language representations to generalized multimodal detection~\cite{mdaa2026}. These methods address rich sensing and fusion settings.

\subsection{Foundation Models and Parameter-Efficient Adaptation}
CLIP pretraining associates visual representations with language across diverse image content~\cite{radford2021learning}. Its visual encoder can subsequently support a supervised PAD head even when the downstream system does not use text prompts. Low-rank adaptation (LoRA) restricts trainable weight changes to a factored subspace while retaining the pretrained matrices~\cite{lora2022}. FoundPAD demonstrates the relevance of this strategy to presentation attack detection~\cite{ozgur2025foundpad}. Accordingly, CLIP with LoRA and binary cross-entropy serves as the direct baseline in our experiments.

Language-guided PAD methods exploit a different part of the pretrained model. FLIP grounds visual representations using language~\cite{srivatsan2023flip}. ME-FAS enhances multimodal textual information for cross-domain PAD~\cite{cai2025mefas}. CLIP-SA uses textual semantics to guide feature alignment~\cite{clipsa2026}; MoGA-ETA develops enhanced text guidance and alignment~\cite{moga2026}. DGPDL models domain signals through a learnable prompt distribution~\cite{dgpdl2026}. These papers establish that pretrained semantic structure can be adapted through text, prompt construction, or visual updates.

\subsection{Masked Modeling and Joint-Embedding Prediction}
Masked autoencoders learn representations by reconstructing withheld image content~\cite{he2022masked}. Bootstrap-style learning uses an online network and an EMA target network to establish representation consistency~\cite{byol2020}. I-JEPA predicts target-region representations from visible context in embedding space~\cite{ijepa2023}. Its relevance here is the predictive relationship between encoded views: the teacher supplies a complete-image target, and the student must infer that target without observing the corresponding image pixels.

Masked and self-predictive ideas already appear in face security. FSFM develops facial representations through self-supervised learning for generalizable face security~\cite{wang2025fsfm}. AdG-MAE combines masked reconstruction, gated attention, and spectral reweighting for domain-generalized PAD~\cite{adgmae2026}. Autoencoders have also been studied for open-set presentation attack detection~\cite{autoenc2026}. In the adjacent digital-forgery setting, structural consistency and self-predictive reconstruction are used to expose facial inconsistencies~\cite{struct2026}. These studies establish masked learning as an existing direction in face-security representation learning.

CP-JEPA instead applies masked latent prediction during supervised low-rank adaptation of CLIP. Its EMA targets are task-adapted representations rather than reconstructed pixels.

\section{Proposed CP-JEPA Framework}
\label{sec:method}

\begin{figure*}[!tb]
    \centering
    \includegraphics[width=0.8\textwidth]{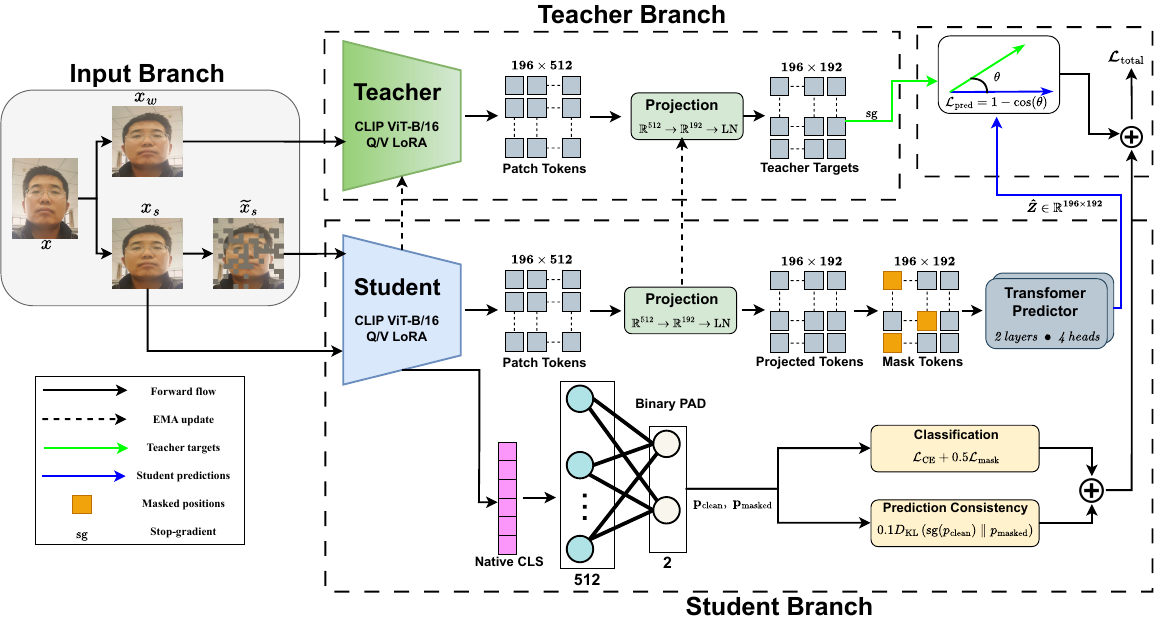}
    \caption{Overview of CP-JEPA training and inference. The student learns from masked spatial predictions and binary classification, using an EMA teacher to provide latent targets. The teacher and predictor are used only during training; inference uses the adapted CLIP encoder and its native binary classifier.}
    \label{fig:framework}
\end{figure*}

CP-JEPA adapts a pretrained CLIP visual encoder for face presentation attack detection (PAD) by combining binary classification with teacher-guided prediction of masked spatial representations. A cross-entropy stage first adapts the encoder and classifier. The subsequent predictive stage retains classification supervision while introducing masked latent prediction, masked classification, and prediction consistency. Fig.~\ref{fig:framework} illustrates the CP-JEPA training architecture.

\subsection{CLIP-LoRA for Face Presentation Attack Detection}

Let $\mathcal{D}_S=\bigcup_{k=1}^{K}\mathcal{D}_k$ denote the source training datasets, with samples $(x,y)$, where $x$ is a face image and $y\in\{0,1\}$ indicates spoof or bona fide, respectively. We use the pretrained CLIP ViT-B/16 visual encoder, whose native global representation is obtained from the class token after the final transformer normalization and visual projection. Let $g_\Delta(x)$ denote this 512-dimensional representation. The binary classifier estimates
\begin{equation}
    p_\theta(y\mid x)
    =\operatorname{softmax}
    \left(C_\omega g_\Delta(x)+b_\omega\right)_y,
    \label{eq:classifier}
\end{equation}
where $\Delta$ denotes the trainable encoder adapters, $\omega$ denotes the classifier parameters, and $\theta=(\Delta,\omega)$. The pretrained CLIP weights remain frozen; the classifier uses the native global representation.

The visual encoder is adapted using low-rank adaptation (LoRA). Trainable low-rank updates are applied to the query and value projections in all 12 transformer blocks. For $U\in\{Q,V\}$, the adapted projection matrix is
\begin{equation}
    W_U=W_{U,0}+\frac{\alpha}{r}B_UA_U,
    \label{eq:lora}
\end{equation}
where $W_{U,0}$ is the frozen pretrained matrix, $A_U\in\R^{r\times d}$ and $B_U\in\R^{d\times r}$ are low-rank factors, and $d=768$. We use $r=8$ and $\alpha=8$. The factors $A_U$ are initialized using Kaiming uniform initialization, while $B_U$ is initialized to zero, leaving the pretrained projections unchanged at initialization. The key projections and remaining pretrained CLIP parameters are frozen. Since the adapters modify attention operations shared by the class token and patch tokens, their updates affect both the global classification representation and the spatial features used for latent prediction.

The first training stage optimizes the adapters and binary classifier using cross-entropy loss:
\begin{equation}
    \CE
    =-\mathbb{E}_{(x,y)\sim\mathcal{D}_S}
    \log p_\theta(y\mid a(x)),
    \label{eq:ce}
\end{equation}
where $a$ denotes the training augmentation. The resulting PAD-adapted model initializes the subsequent predictive training stage, during which classification supervision remains active.

\subsection{Input Views and Patch Masking}

For each training image, two views are generated using a shared geometric transformation $a_g$. The teacher receives $x_w=a_g(x)$, while the student receives a perturbed view $x_s=a_p(a_g(x))$, where $a_p$ applies mild color jitter. The geometric transformation includes resizing and horizontal flipping. Applying the same geometry to both views preserves correspondence between their spatial patch positions.

To construct the masked student input, we randomly select patch positions from the perturbed view. At a resolution of $224\times224$, CLIP ViT-B/16 produces a $14\times14$ grid of $N=196$ spatial tokens. A mask set $\mathcal{M}\subset\{1,\ldots,N\}$ contains $k=\operatorname{round}(\rho N)$ uniformly sampled positions, with $\rho=0.30$. The corresponding image patches are set to zero before student encoding:
\begin{equation}
    \widetilde{x}_s
    =(1-M^\uparrow)\odot x_s,
    \label{eq:inputmask}
\end{equation}
where $M^\uparrow$ denotes the patch mask expanded to image resolution by nearest-neighbor interpolation, and $\odot$ is element-wise multiplication. Masked pixels are set to zero in normalized coordinates, corresponding to the CLIP channel means in the original image space. Applying the mask before the encoder prevents the original masked pixel values from entering the student representation. The predictor subsequently replaces the features at these positions with learned mask tokens.

\subsection{Teacher-Guided Masked Latent Prediction}

Masked latent prediction uses spatial representations from the CLIP visual encoder. Let $H_\Delta(x)\in\R^{N\times768}$ denote the adapted encoder's patch representations after excluding the class token. These features undergo CLIP's post-transformer layer normalization and visual projection, followed by a trainable linear projector $P_\psi:\R^{512}\rightarrow\R^{192}$ and layer normalization. The resulting representation at position $i$ is
\begin{equation}
    z_{\Delta,\psi,i}(x)
    =\operatorname{LN}_{\psi}\left(
    P_\psi\left(
    \operatorname{LN}(H_{\Delta,i}(x))
    W_{\mathrm{proj}}
    \right)
    \right),
    \label{eq:tokens}
\end{equation}
where $\operatorname{LN}$ and $W_{\mathrm{proj}}$ are the pretrained CLIP layer normalization and visual projection, respectively, and $\operatorname{LN}_{\psi}$ is the trainable layer normalization following the spatial projector.

The teacher is initialized from the CE-adapted student encoder and a copy of its newly initialized spatial projector. Its LoRA parameters are denoted by $\bar\Delta$, while $\bar\psi$ denotes the spatial projector and its layer normalization parameters. For the unmasked view $x_w$, the teacher produces normalized spatial targets:
\begin{equation}
    t_i
    =\operatorname{sg}\left(
    \frac{z_{\bar\Delta,\bar\psi,i}(x_w)}
    {\|z_{\bar\Delta,\bar\psi,i}(x_w)\|_2}
    \right),
    \label{eq:target}
\end{equation}
where $\operatorname{sg}$ denotes stop-gradient. The teacher's LoRA and spatial projection parameters are updated after each student optimization step using an exponential moving average (EMA):
\begin{equation}
    \begin{aligned}
    \bar\Delta &\leftarrow
    \mu\bar\Delta+(1-\mu)\Delta,\\
    \bar\psi &\leftarrow
    \mu\bar\psi+(1-\mu)\psi,
    \end{aligned}
    \qquad \mu=0.99.
    \label{eq:ema}
\end{equation}
The pretrained CLIP weights remain fixed in both networks, and the teacher's binary classifier is not used to construct the spatial targets.

The student encodes the masked view $\widetilde{x}_s$ and obtains the projected spatial sequence $Z=z_{\Delta,\psi}(\widetilde{x}_s)$. Before prediction, features at masked positions are replaced with a learned mask token $m$, while visible positions retain their projected features. A fixed two-dimensional sinusoidal positional encoding $e_i$ is added at each position:
\begin{equation}
    u_i=
    \begin{cases}
        m+e_i, & i\in\mathcal{M},\\
        Z_i+e_i, & i\notin\mathcal{M}.
    \end{cases}
    \label{eq:predinput}
\end{equation}

The sequence $U=\{u_i\}_{i=1}^{N}$ is processed by a transformer predictor $q_\phi$ with two layers, four attention heads, a hidden dimension of 192, and a feed-forward dimension of 768. The predictor uses GELU activations and zero dropout. Its output at position $i$ is $\widehat z_i=q_\phi(U)_i$. The prediction loss is the mean cosine distance between the predicted representations and the teacher targets at masked positions:
\begin{equation}
    \LP
    =\frac{1}{|\mathcal{M}|}
    \sum_{i\in\mathcal{M}}
    \left(
    1-\frac{\widehat z_i^{\mathsf T}t_i}
    {\|\widehat z_i\|_2}
    \right).
    \label{eq:predloss}
\end{equation}

The objective predicts masked spatial representations in feature space rather than reconstructing RGB values. It follows the joint-embedding prediction principle of I-JEPA~\cite{ijepa2023}, here incorporated into a supervised PAD model initialized through cross-entropy training.

\subsection{PAD Classification and Consistency Objectives}

The student also performs binary PAD classification on the unmasked view $x_s$ and its masked counterpart $\widetilde{x}_s$. Both forward passes use the same encoder and native classifier. Ordinary cross-entropy supervision is retained for the unmasked view, while the masked view introduces an additional classification loss:
\begin{equation}
    \LM
    =-\mathbb{E}_{(x,y)}
    \log p_\theta(y\mid\widetilde{x}_s).
    \label{eq:maskedce}
\end{equation}

A consistency objective relates the classification distributions produced by the two views. The unmasked prediction serves as a detached reference, while the masked prediction is optimized to match it:
\begin{equation}
    \LC
    =\mathbb{E}_x
    D_{\mathrm{KL}}\left(
    \operatorname{sg}[p_\theta(\cdot\mid x_s)]
    \,\Vert\,
    p_\theta(\cdot\mid\widetilde{x}_s)
    \right).
    \label{eq:consistency}
\end{equation}
The unmasked distribution is detached; the consistency loss backpropagates through the masked-view prediction and its shared encoder and classifier parameters.

The complete predictive training objective combines unmasked classification, masked latent prediction, masked classification, and prediction consistency:
\begin{equation}
    \mathcal{L}
    =\CE+\lambda_p\LP
    +\lambda_m\LM+\lambda_c\LC,
    \label{eq:total}
\end{equation}
with $(\lambda_p,\lambda_m,\lambda_c)=(1,0.5,0.1)$. All auxiliary terms are active from the beginning of predictive adaptation. The prediction loss updates the student adapters through the spatial projector and transformer predictor, whereas the classification and consistency losses operate on the native global predictions. The objectives therefore share the adapted visual encoder while retaining separate spatial prediction and classification pathways.

\subsection{Training and Inference}

The predictive stage is initialized from the cross-entropy stage-transition checkpoint. The student adapters, classifier, spatial projector, and predictor are optimized using~\eqref{eq:total}; the teacher adapters and spatial projection parameters follow the EMA update in~\eqref{eq:ema}.

At inference, classification uses the adapted student encoder and its native binary classifier, following~\eqref{eq:classifier}. The EMA teacher, masking procedure, and transformer predictor are not used. The spatial projector does not contribute to the classification logits. Predictive adaptation optimizes 1,321,410 student-side parameters, of which 295,938 belong to the LoRA adapters and classifier retained at inference. These counts exclude the frozen CLIP encoder.

\section{Experimental Setup}
\label{sec:protocol}

\subsection{Datasets and Evaluation Protocols}

We evaluate \method{} under cross-dataset and unseen-attack PAD protocols. The cross-dataset experiments use OULU-NPU (O)~\cite{boulkenafet2017oulu}, CASIA-MFSD (C)~\cite{zhang2012face}, Idiap Replay-Attack (I)~\cite{chingovska2012effectiveness}, MSU-MFSD (M)~\cite{wen2015face}, and CelebA-Spoof~\cite{zhang2020celeba}. The standard OCIM protocol trains on three datasets and evaluates on the fourth, covering OCI$\rightarrow$M, OMI$\rightarrow$C, OCM$\rightarrow$I, and ICM$\rightarrow$O. The held-out dataset supplies no samples for gradient-based training.

{Single-source evaluation trains on each OCIM dataset individually and tests on the remaining three, producing twelve source--target combinations (Table~\ref{tab:single_source}). Three further configurations examine the effect of training-source composition.} First, CelebA-Spoof is added to each three-source OCIM training pool, with the remaining OCIM dataset held out for evaluation. Second, O+C+M are used for training and CelebA-Spoof for testing. Third, training uses only M+I, with C and O evaluated separately. These configurations examine additional real-source training data, transfer to CelebA-Spoof, and reduced-source training, respectively.

We also evaluate unseen attack categories using the SiW-Mv2 leave-one-attack-type-out protocol (Table~\ref{tab:siwmv2_hter}), which excludes the target attack category from training. 

\subsection{Preprocessing and Source Sampling}

For OCIM videos, faces are detected using MTCNN\cite{7553523}. The highest-confidence bounding box is expanded by a 30\% margin. Training uses a randomly sampled usable frame from each video, while validation and testing use the middle frame. CelebA-Spoof crops follow its provided bounding-box annotations. Images are resized to $224\times224$ and normalized using CLIP channel statistics. Training applies horizontal flipping and color jitter, with brightness, contrast, saturation, and hue magnitudes of $0.15$, $0.15$, $0.10$, and $0.02$, respectively. During predictive adaptation, the teacher and student views share the same geometric transformation to preserve spatial correspondence.

Experiments incorporating CelebA-Spoof as an additional training source use domain- and class-balanced sampling, following established cross-domain PAD practices~\cite{jia2020single,le2024gacfas}. If $n_{k,c}$ denotes the number of training entries in source domain $k$ and class $c$, each entry receives a sampling weight proportional to $1/n_{k,c}$. Sampling is performed with replacement. The epoch budget is twice the number of OCIM source videos, subject to a batch-size-dependent minimum, preventing the larger CelebA-Spoof dataset from determining the training duration.

\subsection{Training and Model Selection}

Training follows the two-stage procedure in Section~\ref{sec:method}, using the selected CE checkpoint to initialize predictive adaptation. Table~\ref{tab:configuration} summarizes the architecture and optimization settings.

\begin{table}[!tb]
\centering
\caption{Main CP-JEPA architecture and training configuration.}
\label{tab:configuration}
\small
\setlength{\tabcolsep}{3pt}
\renewcommand{\arraystretch}{1.08}
\resizebox{0.7\columnwidth}{!}{\begin{tabular}{@{}p{0.40\columnwidth}p{0.54\columnwidth}@{}}
\toprule
Setting & Configuration \\
\midrule
\multicolumn{2}{@{}l}{\textit{Architecture}} \\
Visual encoder & CLIP ViT-B/16 (frozen base) \\
LoRA & Q/V, all 12 blocks; $r=8$, $\alpha=8$ \\
Spatial projector & Linear, $512\rightarrow192$, followed by LayerNorm \\
Predictor & 2 layers; 4 heads; hidden 192; FFN 768 \\
Patch masking & Uniform random; $30\%$ of 196 patches \\
EMA momentum & $\mu=0.99$ \\
Auxiliary loss weights & $\lambda_p=1$, $\lambda_m=0.5$, $\lambda_c=0.1$ \\
\midrule
\multicolumn{2}{@{}l}{\textit{Optimization}} \\
Input resolution & $224\times224$ \\
Training epochs & 40 CE + 40 predictive \\
Batch size / accumulation & 16 / 2 steps \\
CE optimizer & Adam; learning rate $10^{-4}$ \\
Predictive optimizer & AdamW; learning rate $10^{-4}$ \\
Predictive weight decay & 0.01 \\
Predictive gradient clip & 1.0 \\
\bottomrule
\end{tabular}}
\end{table}

Checkpoints are selected by minimum target-validation CE. Target-specific checkpoints are selected separately when multiple targets are evaluated; experiments requiring a common checkpoint use the equally weighted mean validation loss across targets. O and I use their designated development partitions, while development data for C and M are drawn from their official training partitions. CelebA-Spoof development and test membership follow the supplied manifests. Target-development labels are therefore used for model selection, but not for gradient-based training.

The auxiliary-objective study (Table~\ref{tab:ablation_test}) includes CE continuation to control for additional supervised training. All variants retain unmasked CE. The ``prediction only,'' ``masked CE only,'' and ``consistency only'' variants activate the named auxiliary objective alongside unmasked CE. The complete model uses $(\lambda_p,\lambda_m,\lambda_c)=(1,0.5,0.1)$; each removal experiment sets the corresponding coefficient to zero.





\subsection{Performance Measures}

Let $s(x)=p_\theta(y=1\mid x)$ denote the bona fide score, where $y=0$ represents an attack. A sample is classified as bona fide when $s(x)\geq\tau$, where $\tau$ is the decision threshold selected using development partition and frozen before test evaluation. We report attack presentation classification error rate (APCER), bona fide presentation classification error rate (BPCER), half total error rate (HTER), and area under the ROC curve (AUC). The error rates are defined as
\begin{align}
\mathrm{APCER}(\tau)
&=\Pr(s(x)\geq\tau\mid y=0),\\
\mathrm{BPCER}(\tau)
&=\Pr(s(x)<\tau\mid y=1),\\
\mathrm{HTER}(\tau)
&=\tfrac{1}{2}\bigl[
\mathrm{APCER}(\tau)+\mathrm{BPCER}(\tau)
\bigr].
\label{eq:hter}
\end{align}



\section{Results and Discussion}
\label{sec:results}

\subsection{Cross-Dataset Performance on OCIM}

Table~\ref{tab:cross_domain} compares standard OCIM performance with published PAD methods. \method{} achieves the lowest mean HTER of 3.74\% among the compared methods, followed by CCPE at 4.02\%, and a mean AUC of 98.73\%. On OCI$\rightarrow$M and ICM$\rightarrow$O, \method{} achieves both the lowest HTER and highest AUC. On OMI$\rightarrow$C and OCM$\rightarrow$I, other methods achieve better results.

\begin{table*}[!htbp]
\centering
\caption{Cross-domain Face Presentation Attack Detection results on the OCIM benchmark. Lower HTER and higher AUC indicate better performance. The best results are shown in \textbf{bold}, while the second-best results are \underline{underlined}.}
\label{tab:cross_domain}
\resizebox{0.9\linewidth}{!}{
\begin{tabular}{lcccccccccc}
\toprule
\multirow{2}{*}{Method} &
\multicolumn{2}{c}{OCI$\rightarrow$M} &
\multicolumn{2}{c}{OMI$\rightarrow$C} &
\multicolumn{2}{c}{OCM$\rightarrow$I} &
\multicolumn{2}{c}{ICM$\rightarrow$O} &
\multirow{2}{*}{Avg. HTER(\%)$\downarrow$} &
\multirow{2}{*}{Avg. AUC(\%)$\uparrow$} \\
\cmidrule(lr){2-3}\cmidrule(lr){4-5}\cmidrule(lr){6-7}\cmidrule(lr){8-9}
& HTER $\downarrow$ & AUC $\uparrow$ & HTER $\downarrow$ & AUC $\uparrow$ & HTER $\downarrow$ & AUC $\uparrow$ & HTER $\downarrow$ & AUC $\uparrow$ & & \\
\midrule

CIFAS~\cite{liu2022causal} & 5.95 & 96.32 & 10.66 & 95.30 & 8.50 & 97.24 & 13.17 & 93.44 & 9.57 & 95.58 \\











IADG~\cite{zhou2023iadg} & 5.41 & 98.19 & 8.70 & 96.44 & 10.62 & 94.50 & 8.86 & 97.14 & 8.40 & 96.57 \\

DRDN~\cite{he2023dynamic} &
10.00 & 96.20 &
20.90 & 83.80 &
11.40 & 94.50 &
15.60 & 90.90 &
14.50 & 91.30 \\










DiffFAS-R~\cite{ge2024difffas} & 5.90 & 98.10 & 7.32 & 97.40 & 5.66 & 98.69 & 12.38 & 94.33 & 7.82 & 97.13 \\

DiffFAS-V~\cite{ge2024difffas} & \underline{2.86} & 98.41 & 10.11 & 96.32 & 6.36 & 97.89 & 8.11 & 97.27 & 6.86 & 97.47 \\

GAC-FAS~\cite{le2024gacfas} & 5.00 & 97.56 & 8.20 & 95.16 & \underline{4.29} & 98.87 & 8.60 & 97.16 & 6.52 & 97.19 \\

CA-MoEiT~\cite{liu2024camoeit} & 2.88 & 98.76 & 7.89 & 97.70 & 6.18 & 98.94 & 9.72 & 96.22 & 6.67 & 97.91 \\



FoundPAD (ViT-B) \cite{ozgur2025foundpad} & 20.95 & 89.88 & 4.89 & 98.08 & 10.45 & 95.80 & 6.19 & 98.31 & 10.62 & 95.52 \\

FoundPAD (ViT-L) \cite{ozgur2025foundpad} & 16.90 & 93.18 & 6.00 & 98.72 & 9.90 & 96.07 & 5.87 & 98.41 & 9.67 & 96.60 \\

FreqShortcut ~\cite{cao2025fsvfas} & 5.95 & 98.52 & 7.33 & 97.86 & 5.45 & 98.77 & 10.88 & 96.29 & 7.40 & 97.86 \\

FSFM ViT-B~\cite{wang2025fsfm} & 6.58 & 97.43 & 4.30 & 99.10 & 14.63 & 90.96 & 10.02 & 96.23 & 8.88 & 95.93 \\

CCPE~\cite{guo2026ccpe} & 3.10 & \underline{99.21} & \textbf{1.33} & \underline{99.36} & 6.08 & 94.36 & \underline{5.57} & 98.49 & \underline{4.02} & 97.86 \\

SA-TTDG~\cite{chang2026sattdg} & 5.00 & 98.56 & 4.20 & \textbf{99.48} & \textbf{3.87} & \textbf{99.41} & 8.37 & 97.94 & 5.36 & \textbf{98.85} \\


AdG-MAE~\cite{adgmae2026} & 4.59 & 98.75 & 5.21 & 98.60 & 4.34 & \underline{99.22} & 6.81 & \underline{98.85} & 5.24 & \textbf{98.85} \\

\midrule
\rowcolor{gray!20}
\textbf{CP-JEPA (Ours)} & \textbf{2.50} & \textbf{99.73} & \underline{3.33} & 99.12 & 5.25 & 97.19 & \textbf{3.89} & \textbf{98.87} & \textbf{3.74} & \underline{98.73} \\
\bottomrule
\end{tabular}}

\vspace{2pt}
\parbox{\linewidth}{\footnotesize

}
\end{table*}

\subsection{Single-Source Cross-Dataset Performance}

{Table~\ref{tab:single_source} reports single-source transfer across twelve OCIM source--target combinations. CP-JEPA achieves the lowest mean HTER among the compared methods at 11.36\%, improving on FoundPAD ViT-L's 15.49\% by 4.13 percentage points. It obtains the lowest HTER in eight combinations, including all three targets when trained on I. These results demonstrate competitive cross-dataset transfer without requiring multiple source datasets.}

\begin{table*}[!t]
\centering
\caption{Single-source cross-dataset HTER (\%) on OCIM. Each method is trained on one OCIM source dataset and evaluated separately on the remaining three. Lower HTER is better. The best and second-best results are shown in \textbf{bold} and \underline{underlined}, respectively.}
\label{tab:single_source}

\begingroup

\resizebox{0.95\textwidth}{!}{%
\begin{tabular}{l*{12}{c}c}
\toprule
\multirow{2}{*}{Method} &
\multicolumn{3}{c}{Train C} &
\multicolumn{3}{c}{Train I} &
\multicolumn{3}{c}{Train M} &
\multicolumn{3}{c}{Train O} &
\multirow{2}{*}{Avg.} \\
\cmidrule(lr){2-4}
\cmidrule(lr){5-7}
\cmidrule(lr){8-10}
\cmidrule(lr){11-13}
& I & M & O
& C & M & O
& C & I & O
& C & I & M & \\
\midrule

DRDN-ii~\cite{he2023dynamic}
& 27.10 & \textbf{17.10} & 24.50
& 39.10 & 23.10 & 37.10
& 27.80 & \underline{13.00} & 21.70
& 20.10 & 19.00 & 19.90
& 24.13 \\

CF-PAD~\cite{fang2024face}
& 24.80 & \underline{17.14} & 19.43
& 34.00 & 24.76 & 31.70
& 14.44 & 15.90 & 25.34
& 20.33 & 21.50 & \underline{15.00}
& 22.03 \\

FoundPAD (ViT-B)~\cite{ozgur2025foundpad}
& 16.40 & 24.52 & \underline{15.14}
& 17.00 & \underline{18.57} & \underline{13.38}
& 20.00 & 17.10 & \underline{19.41}
& \underline{7.89} & \underline{8.95} & 23.33
& 16.81 \\

FoundPAD (ViT-L)~\cite{ozgur2025foundpad}
& \underline{14.05} & 21.43 & \textbf{11.00}
& \underline{10.22} & 19.29 & 16.94
& \textbf{12.00} & 14.55 & 20.93
& \textbf{7.22} & 14.40 & 23.81
& \underline{15.49} \\

\midrule

\rowcolor{gray!20}
\textbf{CP-JEPA (Ours)}
& \textbf{13.13} & 20.00 & 20.10
& \textbf{4.63} & \textbf{8.33} & \textbf{10.76}
& \underline{12.96} & \textbf{6.88} & \textbf{15.07}
& 8.15 & \textbf{5.00} & \textbf{11.25}
& \textbf{11.36} \\

\bottomrule
\end{tabular}%
}

\endgroup
\end{table*}

\subsection{Additional Real-Source Training Data}

\begin{figure}[!tb]
\centering
\includegraphics[width=0.4\textwidth]
    {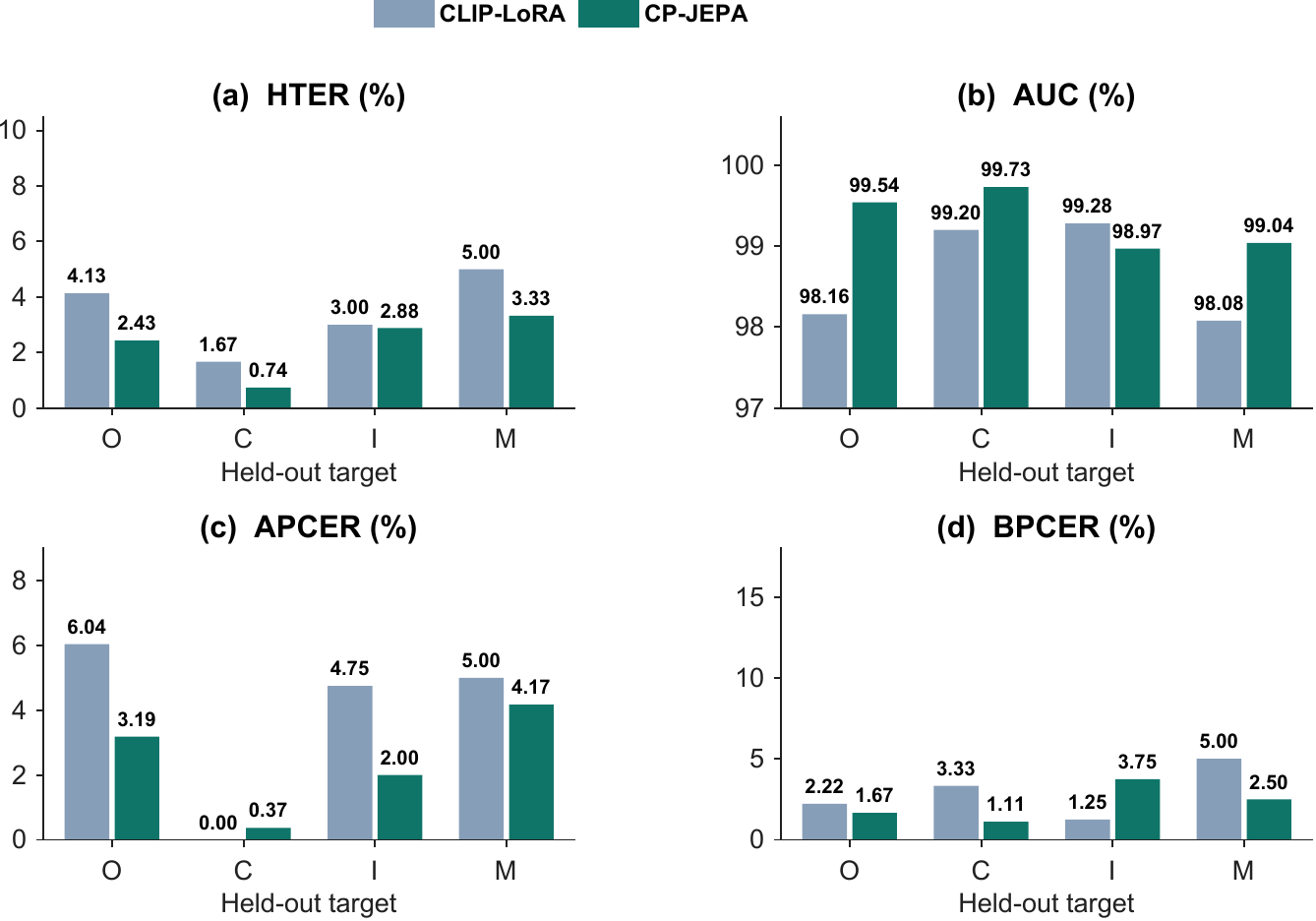}
\caption{Cross-domain HTER, AUC, APCER, and BPCER for CE adaptation and CP-JEPA with CelebA-Spoof added to the source training data.}
\label{fig:cape_performance}
\end{figure}

Adding CelebA-Spoof to the OCIM training sources reduces \method{}'s mean HTER from 3.74\% to 2.35\% and increases AUC from 98.73\% to 99.32\% (Table~\ref{tab:celeb_source}). The improvement in HTER is 1.39 percentage points, or 37.2\% relative to standard OCIM training. Among the listed source-augmented methods, \method{} has the lowest mean HTER, while CCPE achieves a higher mean AUC of 99.55\%. Figure~\ref{fig:cape_performance} compares the source-augmented model with CE adaptation.

\begin{table}[!tb]
\centering
\caption{Performance on the four OCIM leave-one-domain-out protocols with CelebA-Spoof as an additional FacePAD training source. Results are reported as HTER$\downarrow$/AUC$\uparrow$.}
\label{tab:celeb_source}
\resizebox{\linewidth}{!}{%
\begin{tabular}{lccccc}
\toprule
Model &
ICM$\rightarrow$O &
OMI$\rightarrow$C &
OCM$\rightarrow$I &
OCI$\rightarrow$M &
Average \\
\midrule

FLIP-MCL~\cite{srivatsan2023flip}
& \textbf{2.31}/\underline{99.63}
& \textbf{0.54}/\textbf{99.98}
& 4.25/\underline{99.07}
& 4.95/98.11
& 3.01/99.20 \\


FSFM ViT-B~\cite{wang2025fsfm}
& 7.68/97.11
& 3.16/99.41
& 4.63/99.03
& 3.78/\underline{99.15}
& 4.81/98.68 \\

CCPE~\cite{guo2026ccpe}
& 3.64/\textbf{99.67}
& 1.30/\textbf{99.98}
& \underline{4.15}/\textbf{99.31}
& \textbf{2.86}/\textbf{99.24}
& 2.99/\textbf{99.55} \\

\midrule
\rowcolor{gray!20}
\textbf{CP-JEPA (Ours)}
& \underline{2.43}/99.54
& \underline{0.74}/\underline{99.73}
& \textbf{2.88}/98.97
& \underline{3.33}/99.04
& \textbf{2.35}/\underline{99.32} \\

\bottomrule
\end{tabular}%
}
\end{table}

\subsection{Contribution of the Auxiliary Objectives}

Table~\ref{tab:ablation_test} examines the three auxiliary objectives against CE continuation and their individual removal variants. The complete model achieves 3.74\% mean HTER and 98.73\% AUC, compared with 3.97\% and 98.52\% for CE continuation. This comparison controls for additional supervised training after the initial CE stage.

Removing latent prediction raises mean HTER to 3.99\%, while removing masked CE or consistency raises it to 5.05\% and 4.97\%, respectively. The prediction objective does not improve every target: the variant without prediction performs better on O and ties the complete model in HTER on M. 
The complete model achieves the lowest mean HTER among the ablation variants, although the effects differ across targets.

\begin{table}[!tb]
\centering
\caption{Ablation study of the auxiliary objectives in CP-JEPA across the four OCIM protocols.}
\label{tab:ablation_test}
\begingroup
\footnotesize
\setlength{\tabcolsep}{4pt}
\renewcommand{\arraystretch}{1.12}

\resizebox{0.8\linewidth}{!}{\begin{tabular*}{\columnwidth}{@{\extracolsep{\fill}}lrrrrr@{}}
\toprule
Model & O & C & I & M & Avg. \\
\midrule
\multicolumn{6}{@{}l}{\textit{HTER (\%) $\downarrow$}} \\
\midrule
CE
& 3.65 & \textbf{3.33} & \textbf{4.75}
& 4.17 & \underline{3.97} \\
\rowcolor{gray!20}
\textbf{Full}
& 3.89 & \textbf{3.33} & 5.25
& \textbf{2.50} & \textbf{3.74} \\
$-$Pred
& \textbf{3.19} & 5.00 & 5.25
& \textbf{2.50} & 3.99 \\
$-$Mask
& 4.31 & 4.81 & 5.25
& 5.83 & 5.05 \\
$-$Cons
& 4.31 & \underline{3.52} & 5.38
& 6.67 & 4.97 \\
Pred only
& 6.70 & \textbf{3.33} & \underline{4.88}
& 5.00 & 4.98 \\
Mask only
& \underline{3.61} & 6.67 & 5.38
& \underline{3.75} & 4.85 \\
Cons only
& 3.99 & 6.67 & 5.25
& 5.42 & 5.33 \\

\midrule
\multicolumn{6}{@{}l}{\textit{AUC (\%) $\uparrow$}} \\
\midrule
CE
& 98.49 & \textbf{99.30} & \textbf{97.30}
& 98.98 & 98.52 \\
\rowcolor{gray!20}
\textbf{Full}
& \underline{98.87} & \underline{99.12} & 97.19
& \textbf{99.73} & \textbf{98.73} \\
$-$Pred
& \textbf{99.09} & 98.44 & 97.21
& \underline{99.52} & \underline{98.57} \\
$-$Mask
& 98.69 & 98.41 & 97.23
& 98.54 & 98.22 \\
$-$Cons
& 98.70 & \underline{99.12} & 97.17
& 98.58 & 98.39 \\
Pred only
& 95.75 & 99.09 & \underline{97.25}
& 99.04 & 97.78 \\
Mask only
& \textbf{99.09} & 97.27 & 97.21
& 99.33 & 98.22 \\
Cons only
& 98.81 & 97.67 & 97.22
& 98.79 & 98.12 \\
\bottomrule
\end{tabular*}}

\par\vspace{2pt}
{\raggedright\scriptsize
CE: CLIP-LoRA (CE continuation);
Full: CP-JEPA (complete).
Pred: prediction; Mask: masked CE;
Cons: consistency.\par}

\endgroup
\end{table}

\subsection{Generalization to Unseen Attack Categories}
\begin{table*}[!t]
\centering
\caption{HTER (\%) on the SiW-Mv2 leave-one-out protocol. Lower values indicate better performance. The best and second-best results for each attack type are highlighted in \textbf{bold} and \underline{underlined}, respectively.}
\label{tab:siwmv2_hter}
\scriptsize
\resizebox{0.8\textwidth}{!}{
\begin{tabular}{l|cccccccc}
\toprule
\textbf{Attack} &
\textbf{MD-FAS (SRENet)} &
\textbf{DINO (ViT-B-14)} &
\textbf{CLIP-B16} &
\textbf{CLIP-B32} &
\textbf{CLIP-L14} &
\textbf{DINO-S14} &
\textbf{DINO-L14} &
\textbf{CP-JEPA} \\
&
\cite{guo2022multi} &
\cite{gonzalez2025foundation} &
\cite{gonzalez2025foundation} &
\cite{gonzalez2025foundation} &
\cite{gonzalez2025foundation} &
\cite{gonzalez2025foundation} &
\cite{gonzalez2025foundation} &
(Ours) \\
\midrule

FunE.
& 29.50 & \textbf{8.52} & 13.46 & 19.24
& \underline{9.57} & 18.76 & 21.62
& \cellcolor{gray!20}11.57 \\

PEye
& \underline{2.70} & \textbf{0.19} & \textbf{0.19}
& \textbf{0.19} & \textbf{0.19} & \textbf{0.19}
& \textbf{0.19}
& \cellcolor{gray!20}\textbf{0.19} \\

PMouth
& 1.10 & \textbf{0.19} & \underline{0.39}
& 3.65 & \textbf{0.19} & \textbf{0.19}
& \textbf{0.19}
& \cellcolor{gray!20}\textbf{0.19} \\

PaperG.
& 11.90 & \textbf{0.19} & 1.62 & 2.47
& \textbf{0.19} & \underline{1.04}
& \textbf{0.19}
& \cellcolor{gray!20}2.30 \\

Ob.
& \textbf{1.30} & 18.16 & 8.60 & 10.14
& 12.80 & 9.76 & \underline{7.68}
& \cellcolor{gray!20}12.49 \\

Impers.
& 24.50 & 0.39 & \textbf{0.19}
& \underline{0.193} & \textbf{0.19}
& \textbf{0.19} & \textbf{0.19}
& \cellcolor{gray!20}\textbf{0.19} \\

Cosmetic
& 10.90 & 18.88 & 11.37 & \underline{6.36}
& 6.94 & 22.93 & 21.00
& \cellcolor{gray!20}\textbf{3.45} \\

HalfM.
& 8.00 & 1.47 & 3.51 & 5.48
& 1.27 & 6.75 & \underline{1.08}
& \cellcolor{gray!20}\textbf{0.19} \\

Silicone
& 9.20 & \underline{0.58} & \underline{0.58} & 1.74
& \textbf{0.19} & 0.39
& 0.77
& \cellcolor{gray!20}\textbf{0.19} \\

TransM.
& \textbf{0.00} & 1.41 & 3.21 & 5.01
& \underline{0.19} & 5.20
& \underline{0.19}
& \cellcolor{gray!20}2.87 \\

Paper
& 0.60 & \textbf{0.00} & \underline{0.19}
& \textbf{0.00} & \textbf{0.00}
& \textbf{0.00} & \textbf{0.00}
& \cellcolor{gray!20}3.76 \\

Mann.
& \underline{4.00} & \textbf{0.19}
& \textbf{0.19} & \textbf{0.19}
& \textbf{0.19} & \textbf{0.19}
& \textbf{0.19}
& \cellcolor{gray!20}\textbf{0.19} \\

Replay
& 17.90 & \underline{7.05} & 16.27 & 18.84
& 13.00 & \textbf{6.73} & 7.56
& \cellcolor{gray!20}9.22 \\

Print
& 9.60 & 13.99 & 11.72 & 13.79
& \underline{9.45} & 18.16 & 11.72
& \cellcolor{gray!20}\textbf{0.19} \\

\midrule

\textbf{Avg.$\pm$Std.}
& 9.40 $\pm$ 8.80
& 5.09 $\pm$ 7.05
& 5.11 $\pm$ 5.87
& 6.24 $\pm$ 6.75
& \underline{3.88 $\pm$ 5.21}
& 6.46 $\pm$ 8.03
& 5.18 $\pm$ 7.78
& \cellcolor{gray!20}\textbf{3.36 $\pm$ 4.28} \\

\bottomrule
\end{tabular}}
\end{table*}
On SiW-Mv2, \method{} achieves the lowest reported mean HTER in Table~\ref{tab:siwmv2_hter}, at 3.36\%, with an attack-category standard deviation of 4.28\%. The next-lowest mean is 3.88\% for CLIP-L14, giving an absolute difference of 0.52 points and a relative reduction of 13.5\%. The corresponding CLIP-B16 and DINO ViT-B/14 means are 5.11\% and 5.09\%, respectively.

The attack-wise results provide a more detailed comparison. \method{} achieves 0.19\% HTER on seven categories, including half masks, silicone masks, and print attacks, and obtains 3.45\% on cosmetic attacks. Performance remains category-dependent: replay and obfuscation produce HTER values of 9.22\% and 12.49\%, respectively. The mean HTER therefore summarizes results that vary considerably across attack categories.

\subsection{Transfer to CelebA-Spoof and Reduced-Source Training}
For O+C+M$\rightarrow$CelebA-Spoof, \method{} obtains 10.74\% HTER and 93.57\% AUC, with 11.29\% APCER and 10.20\% BPCER (Table~\ref{tab:celeb_target}). This is the lowest HTER among the listed methods, improving on SDA-FAS++ by 3.00 points, while its AUC is 0.13 points lower than the 93.70\% reported for SDA-FAS++. The two measures therefore identify different strengths: the proposed method has the lower error at its selected operating point, whereas SDA-FAS++ has slightly higher ROC discrimination.

\begin{table}[!tb]
\centering
\caption{Cross-dataset evaluation on CelebA-Spoof using O, C, and M as training sources.}
\label{tab:celeb_target}
\scriptsize
\resizebox{0.7\columnwidth}{!}{
\begin{tabular}{lcccc}
\toprule
\textbf{Model} &
\textbf{HTER$\downarrow$} &
\textbf{AUC$\uparrow$} \\
\midrule







Panwar et al.~\cite{panwar2021unsupervised}
& 26.10 & 80.00 \\

CIFAS~\cite{liu2022causal}
& 24.60 & 83.20 \\

SourceOnly~\cite{liu2022source}
& 29.70 & 77.50 \\

SDA-FAS~\cite{liu2022source}
& 18.90 & 90.90 \\

SourceOnly++~\cite{liu2024sdafas}
& 23.87 & 83.67 \\

SDA-FAS++~\cite{liu2024sdafas}
& \underline{13.74} & \textbf{93.70} \\

\midrule

\rowcolor{gray!20}
\textbf{CP-JEPA (Ours)}
& \textbf{10.74} & \underline{93.57} \\

\bottomrule
\end{tabular}}
\end{table}

With only M+I as training sources, \method{} achieves 2.59\% HTER on C and 8.06\% on O, giving a mean of 5.32\% (Table~\ref{tab:two_source}). It achieves the lowest mean HTER among the listed methods, although performance differs substantially between the two targets.

\begin{table}[!t]
\centering
\caption{Cross-dataset evaluation with M and I as training sources and C and O as held-out targets.}
\label{tab:two_source}
\resizebox{1\linewidth}{!}{
\begin{tabular}{lcccccc}
\toprule
\multirow{2}{*}{Model} &
\multicolumn{2}{c}{M+I$\rightarrow$C} &
\multicolumn{2}{c}{M+I$\rightarrow$O} &
\multicolumn{2}{c}{Average} \\
\cmidrule(lr){2-3}\cmidrule(lr){4-5}\cmidrule(lr){6-7}
& HTER $\downarrow$ & AUC $\uparrow$
& HTER $\downarrow$ & AUC $\uparrow$
& HTER $\downarrow$ & AUC $\uparrow$ \\
\midrule

SSDG-M~\cite{jia2020single} & 31.89 & 71.29 & 36.01 & 66.88 & 33.95 & 69.09 \\



D$^2$AM~\cite{chen2021generalizable} & 32.65 & 72.04 & 27.70 & 75.36 & 30.18 & 73.70 \\





SSAN-R~\cite{wang2022domain} & 25.56 & 83.89 & 24.44 & 82.86 & 25.00 & 83.38 \\

HFN+MP~\cite{cai2022learning} & 30.89 & 72.48 & 20.94 & 86.71 & 25.92 & 79.60 \\

MADG~\cite{madg2023} & 23.33 & 82.32 & 26.29 & 81.50 & 24.81 & 81.91 \\


DiVT-M~\cite{liao2023divt} & 20.11 & 86.71 & 23.61 & 85.73 & 21.86 & 86.22 \\

DGUA-FAS~\cite{hong2023domain} & 19.22 & 86.81 & 20.05 & 88.75 & 19.64 & 87.78 \\

DSCI~\cite{long2024dsci} & 16.11 & 89.48 & 23.19 & 84.73 & 19.65 & 87.11 \\

TTDG~\cite{zhou2024ttdg} & 17.77 & 86.69 & 17.70 & 90.09 & 17.74 & 88.39 \\

FQ-FAS~\cite{liu2024fqfas} & 19.72 & 84.69 & 17.05 & 90.05 & 18.39 & 87.37 \\



FreqShortcut~\cite{cao2025fsvfas} & 23.33 & 84.99 & 16.16 & \underline{91.72} & 19.75 & 88.36 \\

DCRN~\cite{dcrn2025} & 15.00 & 91.81 & 19.31 & 89.43 & 17.16 & 90.62 \\

SACL~\cite{niu2026sacl} & 15.56 & 90.56 & 19.16 & 89.48 & 17.36 & 90.02 \\

DSCM-FAS~\cite{xing2026dscm} & \underline{13.44} & 91.89 & 18.38 & 91.64 & \underline{15.91} & \underline{91.77} \\

AdG-MAE~\cite{adgmae2026} & 17.10 & \underline{91.97} & \underline{16.12} & 90.69 & 16.61 & 91.33 \\

\midrule
\rowcolor{gray!20}
\textbf{CP-JEPA (Ours)} & \textbf{2.59} & \textbf{98.12} & \textbf{8.06} & \textbf{95.08} & \textbf{5.32} & \textbf{96.60} \\
\bottomrule
\end{tabular}}
\end{table}

\section{Conclusion}
We presented CP-JEPA, a context-predictive low-rank adaptation framework for face presentation attack detection. The framework combines supervised classification with masked latent prediction, masked classification, and prediction consistency to adapt a pretrained CLIP visual encoder. An EMA teacher provides spatial representation targets, while low-rank attention updates connect the predictive objective to the global classification task. The final classification predictions depend on the adapted student encoder and its native binary classifier, without requiring the teacher or transformer predictor.

Under the reported evaluation, CP-JEPA achieves 3.74\% mean HTER and 98.73\% AUC on OCIM. Adding CelebA-Spoof to the training sources improves these results to 2.35\% and 99.32\%, respectively. The objective ablation supports the combined training formulation, which improves mean HTER over CE continuation and each auxiliary-loss removal variant. {Single-source evaluation yields 11.36\% mean HTER across twelve OCIM transfers, with the lowest HTER in eight combinations among the compared methods. Further evaluations achieve 3.36\% mean HTER on unseen SiW-Mv2 attack categories, 10.74\% HTER on CelebA-Spoof, and 5.32\% mean HTER with two training sources.} These findings support context-predictive supervision as an effective complement to discriminative low-rank adaptation for RGB face PAD, without introducing additional inference modules.



\begin{thebibliography}{56}
\providecommand{\natexlab}[1]{#1}
\providecommand{\url}[1]{#1}
\csname url@samestyle\endcsname
\providecommand{\newblock}{\relax}
\providecommand{\bibinfo}[2]{#2}
\providecommand{\BIBentrySTDinterwordspacing}{\spaceskip=0pt\relax}
\providecommand{\BIBentryALTinterwordstretchfactor}{4}
\providecommand{\BIBentryALTinterwordspacing}{\spaceskip=\fontdimen2\font plus
\BIBentryALTinterwordstretchfactor\fontdimen3\font minus \fontdimen4\font\relax}
\providecommand{\BIBforeignlanguage}[2]{{%
\expandafter\ifx\csname l@#1\endcsname\relax
\typeout{** WARNING: IEEEtranN.bst: No hyphenation pattern has been}%
\typeout{** loaded for the language `#1'. Using the pattern for}%
\typeout{** the default language instead.}%
\else
\language=\csname l@#1\endcsname
\fi
#2}}
\providecommand{\BIBdecl}{\relax}
\BIBdecl

\bibitem[Yu et~al.(2020)Yu, Zhao, Wang, Qin, Su, Li, Zhou, and Zhao]{yu2020searching}
Z.~Yu, C.~Zhao, Z.~Wang, Y.~Qin, Z.~Su, X.~Li, F.~Zhou, and G.~Zhao, ``Searching central difference convolutional networks for face anti-spoofing,'' in \emph{Proceedings of the IEEE/CVF conference on computer vision and pattern recognition}, 2020, pp. 5295--5305.

\bibitem[Jia et~al.(2020)Jia, Zhang, Shan, and Chen]{jia2020single}
Y.~Jia, J.~Zhang, S.~Shan, and X.~Chen, ``Single-side domain generalization for face anti-spoofing,'' in \emph{Proceedings of the IEEE/CVF conference on computer vision and pattern recognition}, 2020, pp. 8484--8493.

\bibitem[Sun et~al.(2023)Sun, Liu, Liu, Li, and Chu]{sun2023safas}
Y.~Sun, Y.~Liu, X.~Liu, Y.~Li, and W.-S. Chu, ``Rethinking domain generalization for face anti-spoofing: Separability and alignment,'' in \emph{Proceedings of the IEEE/CVF Conference on Computer Vision and Pattern Recognition}, 2023, pp. 24\,563--24\,574.

\bibitem[Zhou et~al.(2023)Zhou, Zhang, Yao, Lu, Yi, Ding, and Ma]{zhou2023iadg}
Q.~Zhou, K.-Y. Zhang, T.~Yao, X.~Lu, R.~Yi, S.~Ding, and L.~Ma, ``Instance-aware domain generalization for face anti-spoofing,'' in \emph{Proceedings of the IEEE/CVF Conference on Computer Vision and Pattern Recognition}, 2023, pp. 20\,453--20\,463.

\bibitem[Dharmawan et~al.(2026{\natexlab{a}})Dharmawan, Nugroho, and Handoko]{dfnet2026}
D.~A. Dharmawan, A.~S. Nugroho, and S.~A.~P. Handoko, ``{Toward Generalized and Explainable Face Anti-Spoofing via Decomposition-Fusion Networks},'' \emph{IEEE Transactions on Biometrics, Behavior, and Identity Science}, vol.~8, no.~4, pp. 640--651, 2026.

\bibitem[Yadav et~al.(2026)Yadav, Kavati, and Cheruku]{grdx2026}
K.~D.~K. Yadav, I.~Kavati, and R.~Cheruku, ``{Gradient-Reversed Domain-Generalizable Multi-Modal Cross-Attention Network for Robust Face Anti-Spoofing},'' \emph{IEEE Transactions on Biometrics, Behavior, and Identity Science}, vol.~8, no.~1, pp. 84--98, 2026.

\bibitem[Dharmawan et~al.(2026{\natexlab{b}})Dharmawan, Krisna, Nugroho, Ferriyan, Yuwono, Widiwasita, and Hadi]{mbca2026}
D.~A. Dharmawan, M.~V.~A. Krisna, A.~S. Nugroho, A.~Ferriyan, B.~Yuwono, J.~Widiwasita, and A.~J.~S. Hadi, ``{FAS-MBCA: Face Anti-Spoofing Using Multi-Path Bi-Directional Cross-Attention Fusion},'' \emph{IEEE Transactions on Biometrics, Behavior, and Identity Science}, vol.~8, no.~5, pp. 947--959, 2026.

\bibitem[Ozgur et~al.(2025)Ozgur, Caldeira, Chettaoui, Boutros, Ramachandra, and Damer]{ozgur2025foundpad}
G.~Ozgur, E.~Caldeira, T.~Chettaoui, F.~Boutros, R.~Ramachandra, and N.~Damer, ``Foundpad: Foundation models reloaded for face presentation attack detection,'' in \emph{Proceedings of the Winter Conference on Applications of Computer Vision}, 2025, pp. 745--755.

\bibitem[Assran et~al.(2023)Assran, Duval, Misra, Bojanowski, Vincent, Rabbat, LeCun, and Ballas]{ijepa2023}
M.~Assran, Q.~Duval, I.~Misra, P.~Bojanowski, P.~Vincent, M.~Rabbat, Y.~LeCun, and N.~Ballas, ``{Self-Supervised Learning from Images with a Joint-Embedding Predictive Architecture},'' in \emph{2023 IEEE/CVF Conference on Computer Vision and Pattern Recognition (CVPR)}, 2023, pp. 15\,619--15\,629.

\bibitem[He et~al.(2023)He, Peng, and Long]{he2023dynamic}
Y.~He, F.~Peng, and M.~Long, ``Dynamic residual distillation network for face anti-spoofing with feature attention learning,'' \emph{IEEE Transactions on Biometrics, Behavior, and Identity Science}, vol.~5, no.~4, pp. 579--592, 2023.

\bibitem[Ramachandra et~al.(2025)Ramachandra, Vetrekar, Patwardhan, Venkatesh, Naik, and Gad]{ramachandra2025pcgattnnet}
R.~Ramachandra, N.~Vetrekar, S.~Patwardhan, S.~Venkatesh, G.~Naik, and R.~S. Gad, ``{PCGattnNet}: A 3-d point cloud dynamic graph attention for generalizable face presentation attack detection,'' \emph{IEEE Transactions on Biometrics, Behavior, and Identity Science}, vol.~7, no.~4, pp. 924--939, 2025.

\bibitem[Diao et~al.(2026)Diao, Li, Guo, Lin, Liu, Wan, Zhou, Yu, and Lei]{mdaa2026}
Y.~Diao, Y.~Li, J.~Guo, X.~Lin, A.~Liu, J.~Wan, J.~T. Zhou, Z.~Yu, and Z.~Lei, ``{Modality- and Domain-Aware Adaptivity for Generalized Multimodal Face Anti-Spoofing},'' \emph{IEEE Transactions on Information Forensics and Security}, vol.~21, pp. 8963--8978, 2026.

\bibitem[Radford et~al.(2021)Radford, Kim, Hallacy, Ramesh, Goh, Agarwal, Sastry, Askell, Mishkin, Clark, Krueger, and Sutskever]{radford2021learning}
A.~Radford, J.~W. Kim, C.~Hallacy, A.~Ramesh, G.~Goh, S.~Agarwal, G.~Sastry, A.~Askell, P.~Mishkin, J.~Clark, G.~Krueger, and I.~Sutskever, ``Learning transferable visual models from natural language supervision,'' in \emph{Proceedings of the 38th International Conference on Machine Learning}, vol. 139, 2021, pp. 8748--8763.

\bibitem[Hu et~al.(2022)Hu, Shen, Wallis, Allen-Zhu, Li, Wang, Wang, and Chen]{lora2022}
E.~J. Hu, Y.~Shen, P.~Wallis, Z.~Allen-Zhu, Y.~Li, S.~Wang, L.~Wang, and W.~Chen, ``{LoRA: Low-Rank Adaptation of Large Language Models},'' in \emph{International Conference on Learning Representations}, 2022.

\bibitem[Srivatsan et~al.(2023)Srivatsan, Naseer, and Nandakumar]{srivatsan2023flip}
K.~Srivatsan, M.~Naseer, and K.~Nandakumar, ``Flip: Cross-domain face anti-spoofing with language guidance,'' in \emph{Proceedings of the IEEE/CVF international conference on computer vision}, 2023, pp. 19\,685--19\,696.

\bibitem[Cai et~al.(2025)Cai, Wang, Ji, Sun, Cao, and Ji]{cai2025mefas}
L.~Cai, H.~Wang, J.~Ji, X.~Sun, L.~Cao, and R.~Ji, ``{ME-FAS}: Multimodal text enhancement for cross-domain face anti-spoofing,'' \emph{IEEE Transactions on Information Forensics and Security}, vol.~20, 2025.

\bibitem[Ma et~al.(2026)Ma, Liu, Xue, Diao, Liang, Liang, Liang, Wan, Yang, Hu, and Lei]{clipsa2026}
H.~Ma, A.~Liu, S.~Xue, Y.~Diao, D.~Liang, H.~Liang, Y.~Liang, J.~Wan, X.~Yang, Z.~Hu, and Z.~Lei, ``{CLIP-SA: CLIP-Guided Semantic Alignment for Generalizable Face Anti-spoofing},'' \emph{IEEE Transactions on Multimedia}, pp. 1--16, 2026, early access.

\bibitem[Zhang et~al.(2026)Zhang, Liu, Xie, Huang, Yu, and Kang]{moga2026}
L.~Zhang, K.~Liu, T.~Xie, J.~Huang, Z.~Yu, and W.~Kang, ``{MoGA-ETA: Generalized Face Anti-Spoofing With Enhanced Text Guidance and Alignment},'' \emph{IEEE Transactions on Biometrics, Behavior, and Identity Science}, vol.~8, no.~3, pp. 340--353, 2026.

\bibitem[Liu et~al.(2026)Liu, Lin, Zhi, Liang, Zhu, Cai, Wan, Escalera, and Lei]{dgpdl2026}
A.~Liu, X.~Lin, R.~Zhi, Y.~Liang, X.~Zhu, Z.~Cai, J.~Wan, S.~Escalera, and Z.~Lei, ``{DGPDL: Domain-Guided Prompt Distribution Learning for Generalizable Face Anti-Spoofing},'' \emph{IEEE Transactions on Pattern Analysis and Machine Intelligence}, vol.~48, no.~8, pp. 8953--8969, 2026.

\bibitem[He et~al.(2022)He, Chen, Xie, Li, Dollar, and Girshick]{he2022masked}
\BIBentryALTinterwordspacing
K.~He, X.~Chen, S.~Xie, Y.~Li, P.~Dollar, and R.~Girshick, ``Masked autoencoders are scalable vision learners,'' in \emph{Proceedings of the IEEE/CVF Conference on Computer Vision and Pattern Recognition}, 2022, pp. 16\,000--16\,009. [Online]. Available: \url{https://openaccess.thecvf.com/content/CVPR2022/html/He_Masked_Autoencoders_Are_Scalable_Vision_Learners_CVPR_2022_paper.html}
\BIBentrySTDinterwordspacing

\bibitem[Grill et~al.(2020)Grill, Strub, Altch{\'e}, Tallec, Richemond, Buchatskaya, Doersch, Pires, Guo, Azar, Piot, Kavukcuoglu, Munos, and Valko]{byol2020}
J.-B. Grill, F.~Strub, F.~Altch{\'e}, C.~Tallec, P.~H. Richemond, E.~Buchatskaya, C.~Doersch, B.~A. Pires, Z.~D. Guo, M.~G. Azar, B.~Piot, K.~Kavukcuoglu, R.~Munos, and M.~Valko, ``{Bootstrap Your Own Latent: A New Approach to Self-Supervised Learning},'' in \emph{Advances in Neural Information Processing Systems}, vol.~33, 2020, pp. 21\,271--21\,284.

\bibitem[Wang et~al.(2025)Wang, Lin, Wu, Liu, Ba, and Ren]{wang2025fsfm}
G.~Wang, F.~Lin, T.~Wu, Z.~Liu, Z.~Ba, and K.~Ren, ``{FSFM}: A generalizable face security foundation model via self-supervised facial representation learning,'' in \emph{Proceedings of the IEEE/CVF Conference on Computer Vision and Pattern Recognition}, 2025, pp. 24\,364--24\,376.

\bibitem[Sheikhfathollahi et~al.(2026)Sheikhfathollahi, Parkinson, and Khan]{adgmae2026}
M.~Sheikhfathollahi, S.~Parkinson, and S.~Khan, ``{AdG-MAE}: Adaptive gated masked autoencoder for domain-generalised face anti-spoofing,'' \emph{IEEE Transactions on Biometrics, Behavior, and Identity Science}, vol.~8, no.~6, pp. 1211--1224, 2026.

\bibitem[Bastos et~al.(2026)Bastos, George, Marcel, and Rocha]{autoenc2026}
I.~Bastos, A.~George, S.~Marcel, and A.~Rocha, ``{Autoencoders for Open-Set Presentation Attack Detection},'' \emph{IEEE Transactions on Biometrics, Behavior, and Identity Science}, vol.~8, no.~2, pp. 192--206, 2026.

\bibitem[Zhou et~al.(2026)Zhou, Chen, and Ruan]{struct2026}
L.~Zhou, M.~Chen, and H.~Ruan, ``{Structural Consistency for Face Forgery Detection via Frequency Domain Enhancement and Self-Predictive Learning},'' \emph{IEEE Transactions on Biometrics, Behavior, and Identity Science}, vol.~8, no.~1, pp. 99--110, 2026.

\bibitem[Boulkenafet et~al.(2017)Boulkenafet, Komulainen, Li, Feng, and Hadid]{boulkenafet2017oulu}
Z.~Boulkenafet, J.~Komulainen, L.~Li, X.~Feng, and A.~Hadid, ``Oulu-npu: A mobile face presentation attack database with real-world variations,'' in \emph{2017 12th IEEE international conference on automatic face \& gesture recognition (FG 2017)}.\hskip 1em plus 0.5em minus 0.4em\relax IEEE, 2017, pp. 612--618.

\bibitem[Zhang et~al.(2012)Zhang, Yan, Liu, Lei, Yi, and Li]{zhang2012face}
Z.~Zhang, J.~Yan, S.~Liu, Z.~Lei, D.~Yi, and S.~Z. Li, ``A face antispoofing database with diverse attacks,'' in \emph{2012 5th IAPR international conference on Biometrics (ICB)}.\hskip 1em plus 0.5em minus 0.4em\relax IEEE, 2012, pp. 26--31.

\bibitem[Chingovska et~al.(2012)Chingovska, Anjos, and Marcel]{chingovska2012effectiveness}
I.~Chingovska, A.~Anjos, and S.~Marcel, ``On the effectiveness of local binary patterns in face anti-spoofing,'' in \emph{2012 BIOSIG-proceedings of the international conference of biometrics special interest group (BIOSIG)}.\hskip 1em plus 0.5em minus 0.4em\relax IEEE, 2012, pp. 1--7.

\bibitem[Wen et~al.(2015)Wen, Han, and Jain]{wen2015face}
D.~Wen, H.~Han, and A.~K. Jain, ``Face spoof detection with image distortion analysis,'' \emph{IEEE Transactions on Information Forensics and Security}, vol.~10, no.~4, pp. 746--761, 2015.

\bibitem[Zhang et~al.(2020)Zhang, Yin, Li, Yin, Yan, Shao, and Liu]{zhang2020celeba}
Y.~Zhang, Z.~Yin, Y.~Li, G.~Yin, J.~Yan, J.~Shao, and Z.~Liu, ``Celeba-spoof: Large-scale face anti-spoofing dataset with rich annotations,'' in \emph{European conference on computer vision}.\hskip 1em plus 0.5em minus 0.4em\relax Springer, 2020, pp. 70--85.

\bibitem[Zhang et~al.(2016)Zhang, Zhang, Li, and Qiao]{7553523}
K.~Zhang, Z.~Zhang, Z.~Li, and Y.~Qiao, ``Joint face detection and alignment using multitask cascaded convolutional networks,'' \emph{IEEE Signal Processing Letters}, vol.~23, no.~10, pp. 1499--1503, Oct 2016.

\bibitem[Le and Woo(2024)]{le2024gacfas}
B.~M. Le and S.~S. Woo, ``Gradient alignment for cross-domain face anti-spoofing,'' in \emph{Proceedings of the IEEE/CVF Conference on Computer Vision and Pattern Recognition}, 2024.

\bibitem[Liu et~al.(2022{\natexlab{a}})Liu, Chen, Dai, Li, Zou, and Xiong]{liu2022causal}
Y.~Liu, Y.~Chen, W.~Dai, C.~Li, J.~Zou, and H.~Xiong, ``Causal intervention for generalizable face anti-spoofing,'' in \emph{2022 IEEE International Conference on Multimedia and Expo (ICME)}.\hskip 1em plus 0.5em minus 0.4em\relax IEEE, 2022, pp. 01--06.

\bibitem[Ge et~al.(2024)Ge, Liu, Yu, Shi, Qi, Li, and Kalviainen]{ge2024difffas}
X.~Ge, X.~Liu, Z.~Yu, J.~Shi, C.~Qi, J.~Li, and H.~Kalviainen, ``{DiffFAS}: Face anti-spoofing via generative diffusion models,'' in \emph{European Conference on Computer Vision}, 2024.

\bibitem[Liu(2024)]{liu2024camoeit}
A.~Liu, ``Ca-moeit: Generalizable face anti-spoofing via dual cross-attention and semi-fixed mixture-of-expert,'' \emph{International Journal of Computer Vision}, vol. 132, no.~11, pp. 5439--5452, 2024.

\bibitem[Cao and Ma(2025)]{cao2025fsvfas}
J.~Cao and C.~Ma, ``Towards generalized face anti-spoofing from a frequency shortcut view,'' in \emph{Proceedings of the IEEE/CVF Winter Conference on Applications of Computer Vision}, 2025, pp. 1005--1015.

\bibitem[Guo et~al.(2026)Guo, Liu, Diao, Zhang, Ma, Zhao, Hong, and Wang]{guo2026ccpe}
J.~Guo, A.~Liu, Y.~Diao, J.~Zhang, H.~Ma, B.~Zhao, R.~Hong, and M.~Wang, ``Domain generalization for face anti-spoofing via content-aware composite prompt engineering,'' \emph{IEEE Transactions on Multimedia}, vol.~28, 2026.

\bibitem[Chang et~al.(2026)Chang, Shi, and Zhang]{chang2026sattdg}
X.~Chang, L.~Shi, and A.~Zhang, ``Semantically anchored test-time domain generalization for face anti-spoofing,'' \emph{Computers, Materials \& Continua}, vol.~89, no.~1, 2026.

\bibitem[Fang and Damer(2024)]{fang2024face}
M.~Fang and N.~Damer, ``Face presentation attack detection by excavating causal clues and adapting embedding statistics,'' in \emph{Proceedings of the IEEE/CVF Winter Conference on Applications of Computer Vision}, 2024, pp. 6269--6279.

\bibitem[Guo et~al.(2022)Guo, Liu, Jain, and Liu]{guo2022multi}
X.~Guo, Y.~Liu, A.~Jain, and X.~Liu, ``Multi-domain learning for updating face anti-spoofing models,'' in \emph{European conference on computer vision}.\hskip 1em plus 0.5em minus 0.4em\relax Springer, 2022, pp. 230--249.

\bibitem[Gonzalez-Soler et~al.(2025)Gonzalez-Soler, Tapia, and Busch]{gonzalez2025foundation}
L.~J. Gonzalez-Soler, J.~E. Tapia, and C.~Busch, ``Are foundation models all you need for zero-shot face presentation attack detection?'' in \emph{2025 IEEE 19th International Conference on Automatic Face and Gesture Recognition (FG)}.\hskip 1em plus 0.5em minus 0.4em\relax IEEE, 2025, pp. 1--10.

\bibitem[Panwar et~al.(2021)Panwar, Singh, Saha, Paudel, and Van~Gool]{panwar2021unsupervised}
A.~Panwar, P.~Singh, S.~Saha, D.~P. Paudel, and L.~Van~Gool, ``Unsupervised compound domain adaptation for face anti-spoofing,'' in \emph{2021 16th IEEE international conference on automatic face and gesture recognition (FG 2021)}.\hskip 1em plus 0.5em minus 0.4em\relax IEEE, 2021, pp. 1--8.

\bibitem[Liu et~al.(2022{\natexlab{b}})Liu, Chen, Dai, Gou, Huang, and Xiong]{liu2022source}
Y.~Liu, Y.~Chen, W.~Dai, M.~Gou, C.-T. Huang, and H.~Xiong, ``Source-free domain adaptation with contrastive domain alignment and self-supervised exploration for face anti-spoofing,'' in \emph{Computer Vision -- ECCV 2022}.\hskip 1em plus 0.5em minus 0.4em\relax Springer, 2022, pp. 511--528.

\bibitem[Liu et~al.(2024{\natexlab{a}})Liu, Chen, Dai, Gou, Huang, and Xiong]{liu2024sdafas}
------, ``Source-free domain adaptation with domain generalized pretraining for face anti-spoofing,'' \emph{IEEE Transactions on Pattern Analysis and Machine Intelligence}, vol.~46, no.~8, pp. 5430--5448, 2024.

\bibitem[Chen et~al.(2021)Chen, Yao, Sheng, Ding, Tai, Li, Huang, and Jin]{chen2021generalizable}
Z.~Chen, T.~Yao, K.~Sheng, S.~Ding, Y.~Tai, J.~Li, F.~Huang, and X.~Jin, ``Generalizable representation learning for mixture domain face anti-spoofing,'' in \emph{Proceedings of the AAAI conference on artificial intelligence}, vol.~35, no.~2, 2021, pp. 1132--1139.

\bibitem[Wang et~al.(2022)Wang, Wang, Yu, Deng, Li, Gao, and Wang]{wang2022domain}
Z.~Wang, Z.~Wang, Z.~Yu, W.~Deng, J.~Li, T.~Gao, and Z.~Wang, ``Domain generalization via shuffled style assembly for face anti-spoofing,'' in \emph{2022 IEEE/CVF Conference on Computer Vision and Pattern Recognition (CVPR)}.\hskip 1em plus 0.5em minus 0.4em\relax IEEE, 2022, pp. 4113--4123.

\bibitem[Cai et~al.(2022)Cai, Li, Wan, Li, Hu, and Kot]{cai2022learning}
R.~Cai, Z.~Li, R.~Wan, H.~Li, Y.~Hu, and A.~C. Kot, ``Learning meta pattern for face anti-spoofing,'' \emph{IEEE Transactions on Information Forensics and Security}, vol.~17, pp. 1201--1213, 2022.

\bibitem[Zhang and Nie(2023)]{madg2023}
S.~Zhang and W.~Nie, ``Multi-domain feature alignment for face anti-spoofing,'' \emph{Sensors}, vol.~23, no.~8, p. 4077, 2023.

\bibitem[Liao et~al.(2023)Liao, Chen, Liu, Yeh, Hu, and Chen]{liao2023divt}
C.-H. Liao, W.-C. Chen, H.-T. Liu, Y.-R. Yeh, M.-C. Hu, and C.-S. Chen, ``Domain invariant vision transformer learning for face anti-spoofing,'' in \emph{Proceedings of the IEEE/CVF Winter Conference on Applications of Computer Vision}, 2023, pp. 6098--6107.

\bibitem[Hong et~al.(2023)Hong, Lin, Liu, Yeh, and Chen]{hong2023domain}
Z.-W. Hong, Y.-C. Lin, H.-T. Liu, Y.-R. Yeh, and C.-S. Chen, ``Domain-generalized face anti-spoofing with unknown attacks,'' in \emph{2023 IEEE International Conference on Image Processing}, 2023, pp. 820--824.

\bibitem[Long et~al.(2024)Long, Zhang, Wu, Jin, and Shan]{long2024dsci}
X.~Long, J.~Zhang, S.~Wu, X.~Jin, and S.~Shan, ``Dual sampling based causal intervention for face anti-spoofing with identity debiasing,'' \emph{IEEE Transactions on Information Forensics and Security}, vol.~19, pp. 851--862, 2024.

\bibitem[Zhou et~al.(2024)Zhou, Zhang, Yao, Lu, Ding, and Ma]{zhou2024ttdg}
Q.~Zhou, K.-Y. Zhang, T.~Yao, X.~Lu, S.~Ding, and L.~Ma, ``Test-time domain generalization for face anti-spoofing,'' in \emph{Proceedings of the IEEE/CVF Conference on Computer Vision and Pattern Recognition}, 2024, pp. 175--187.

\bibitem[Liu et~al.(2024{\natexlab{b}})Liu, Li, Li, Wang, and Wu]{liu2024fqfas}
Y.~Liu, Z.~Li, S.~Li, Z.~Wang, and L.~Wu, ``Face anti-spoofing via interaction learning with face image quality alignment,'' in \emph{2024 IEEE 18th International Conference on Automatic Face and Gesture Recognition (FG)}, 2024, pp. 1--10.

\bibitem[Liu et~al.(2025)Liu, Li, and Wu]{dcrn2025}
Y.~Liu, Z.~Li, and L.~Wu, ``Dual consistency regularization for generalized face anti-spoofing,'' \emph{IEEE Transactions on Information Forensics and Security}, vol.~20, pp. 2171--2183, 2025.

\bibitem[Niu and Lin(2026)]{niu2026sacl}
Y.~Niu and X.~Lin, ``Similarity-aware contrastive learning for face anti-spoofing via frequency enhancement and reconstruction,'' \emph{Neural Networks}, vol. 199, p. 108734, 2026.

\bibitem[Xing et~al.(2026)Xing, Tan, Qamar, and Jiao]{xing2026dscm}
H.~Xing, S.~Y. Tan, F.~Qamar, and Y.~Jiao, ``Dual semantic consistency module for domain generalizable face anti-spoofing,'' \emph{Journal of King Saud University - Computer and Information Sciences}, vol.~38, p. 829, 2026.

\end{thebibliography}
\end{document}